\documentclass{article}
\usepackage{dingbat}
\usepackage{spconf,amsmath,graphicx}
\usepackage{tabularx}
\usepackage{xcolor}
\usepackage{flushend}
\usepackage{verbatim}
\usepackage{pgfplots}
\pgfplotsset{width=7cm,compat=1.8}
\usepackage{pgfplotstable}
\usepackage{enumitem}

\usepackage{sfmath}

\title{ActioNet: An interactive end-to-end platform for task-based data collection and augmentation in 3D environment}
\name{Jiafei Duan$^{\star}$ \qquad Samson Yu$^{\star}$ \qquad Hui Li Tan$^{\dagger}$ \qquad Cheston Tan$^{\dagger\star}$
\thanks{Jiafei Duan and Samson Yu performed the work while at A*STAR.}  
}
\address{duan0038@e.ntu.edu.sg, samson$\_$yu@mymail.sutd.edu.sg,  \{hltan, cheston-tan\}@i2r.a-star.edu.sg \\
$^{\star}$ A*STAR Artificial Intelligence Initiative, Singapore \\
  $^{\dagger}$Institute for Infocomm Research, A*STAR}
\begin{document}
%
\maketitle
\begin{abstract}
The problem of task planning for artificial agents remains largely unsolved. 
While there has been increasing interest in data-driven approaches for the study of task planning for artificial agents, a significant remaining bottleneck is the dearth of large-scale comprehensive task-based datasets.
In this paper, we present ActioNet, an interactive end-to-end platform for data collection and augmentation of task-based dataset in 3D environment.
Using ActioNet, we collected a large-scale comprehensive task-based dataset, comprising over 3000 hierarchical task structures and videos. 
Using the hierarchical task structures, the videos are further augmented across 50 different scenes to give over 150,000 video.
To our knowledge, ActioNet is the first interactive end-to-end platform for such task-based dataset generation and the accompanying dataset is the largest task-based dataset of such comprehensive nature.
The ActioNet platform and dataset will be made available to facilitate research in hierarchical task planning. 
The source code, platform, and dataset will be made available\footnote{https://github.com/SamsonYuBaiJian/actionet}.
\end{abstract}
\begin{keywords}
hierarchical task planning, artificial agent, platform, dataset, 3D environment
\end{keywords}
\section{Introduction}
\label{sec:introduction}
 Recent advancements in robotics such as Boston Dynamics's Atlas or social robot Sophia are pushing new frontiers in social robots that exist in our daily life. 
Despite the advancements, existing robots are still limited in their task planning capabilities. 
For instance, when human are instructed to complete a task such as ``make a cup of coffee", they can naturally decompose the task into its sub-tasks and complete its sub-tasks progressively, with little prior knowledge of the task and environment. 
On the other hand, while existing robots can perform pre-learned tasks in pre-learned scenes, they can hardly generalize well to new tasks and scenes. 
Besides reinforcement learning \cite{Wortsman_2019_CVPR}, there is increasing interest in data-driven approach \cite{gordon2018iqa} \cite{Yang2018VisualSN} \cite{7989381} for the study of task planning for artificial agents. 
Nonetheless, while there have been recent developments of photo-realistic 3D environment for the collection of task-based datasets in 3D environment, existing task-based datasets remain limited in scale and comprehensiveness. 

We present ActioNet, an interactive end-to-end platform for task-based dataset collection and augmentation in 3D environment. 
To our knowledge, ActioNet is the first platform enabling large-scale comprehensive task-based data collection and augmentation in 3D environment with minimal human intervention. 
ActioNet allows the collection of video and hierarchical task structure data from human demonstration in 3D environment. 
Leveraging on AI2THOR \cite{ai2thor}, the state-of-the-art interactive simulation platform in 3D home environment, 
ActioNet offers a wide coverage of scenes and settings, as well as interactive control of the artificial agent for real-world physics-based video data generation.
On top of the video data, the hierarchical task structure data provides detailed task breakdown, from high-level goal to second-level task descriptions as performed by human to third-level atomic/robotic actions.
Using the hierarchical task structure, ActioNet allows further augmentation of the video data, from its source scene to different scenes.

Using the ActioNet platform, we collected a large-scale comprehensive task-based dataset, comprising over 3000 videos and hierarchical task structures data over 65 household tasks across 120 scenes.  
Using the hierarchical task structures, the videos are further augmented across 50 scenes, giving a total of over 150,000 videos. 
We believe that this large-scale comprehensive task-based dataset will facilitate research in hierarchical task planning, and is a key to building better hierarchical task-oriented learning models for artificial agents. 
Moreover, ActioNet will be made available as an open-source platform to promote further task-based data collection and augmentation.

The related works will be covered in Section \ref{sec:relatedworks}.
The ActioNet platform and dataset will be elaborated in Sections \ref{sec:actionnetplatform} and \ref{sec:actionetdataset} respectively.
Finally, the conclusion will be presented in Section \ref{sec:conclusion}.

\section{Related works}
\label{sec:relatedworks}
There are few related works for task-based data collection in 3D environment. 
VR Kitchen \cite{VRKitchen} is a 3D environment of virtual kitchens that allows the collection of human cooking demonstration dataset. 
The accompanying VR Chef dataset consists of 20 demonstrations for five different dish preparation tasks. Each demonstration has an average of 25 steps, where each step is annotated via interaction with the virtual environment using VR headset and touch controller. 
On the other hand,
Virtual Home \cite{puig2018virtualhome} is a 3D environment of apartment houses. 
The accompany activity program dataset is collected across six different apartment settings in the Virtual Home simulator, and  consists of 2821 household task program descriptions with many repetitive tasks.
Amazon Mechanical Turk (AMT) workers first provide task descriptions of various household tasks, and then use the task descriptions to further generate program descriptions using MIT's Scratch project \cite{WinNT}. 
Unlike Virtual Home’s program description collection method, the hierarchical task structure of ActioNet is collected by getting users to execute the tasks within the environment via ActioNet’s graphical user interface (GUI). 
This allows users to execute task planning in realistic simulation, hence making task planning more accurate as compared to generating sets of instructions. 


Existing efforts to collect task-based dataset require a lot of human interventions throughout the collection process, and is prohibitive for the collection of large-scale datasets. 
In contrast, ActioNet is an interactive end-to-end collection platform that requires minimal human intervention in both data collection and augmentation. 
Referring to Table \ref{tab:comparisonadvantages}, ActioNet has various advantages over existing collection pipelines.
ActioNet provides a large number of different scenes and settings, and the ability to configure and scale these scene and settings to generate videos of different task scenarios.
ActioNet also provides a physics-based engine that generates videos that realistically simulate real-world interactions.
Furthermore, ActioNet provides interactive control of the artificial agent to generate videos from human demonstration, and such data can be further up-scaled through ActioNet's data augmentation unit thereby allowing imitation learning or multi-modalities deep learning.


\begin{table}[htbp]
  \centering
  \caption{A comparison of task-based data collection pipelines in 3D environment.}
    \begin{tabular}{|l|c|c|c|} \hline
          & VR Kitchen & Virtual Home & ActioNet \\ \hline
    Large-scale &       &       & \checkmark \\ \hline
    Physics-based & \checkmark &       & \checkmark \\ \hline
    Configurable &       & \checkmark & \checkmark \\ \hline
    Scalable &       & \checkmark & \checkmark \\ \hline
    Interactive & \checkmark &  & \checkmark \\ \hline
    \end{tabular}%
  \label{tab:comparisonadvantages}%
\end{table}%

\section{ActioNet platform}
\label{sec:actionnetplatform}
\begin{figure*}[thb]
\begin{minipage}[b]{1.0\linewidth}
  \centering
  \centerline{\includegraphics[width=18cm]{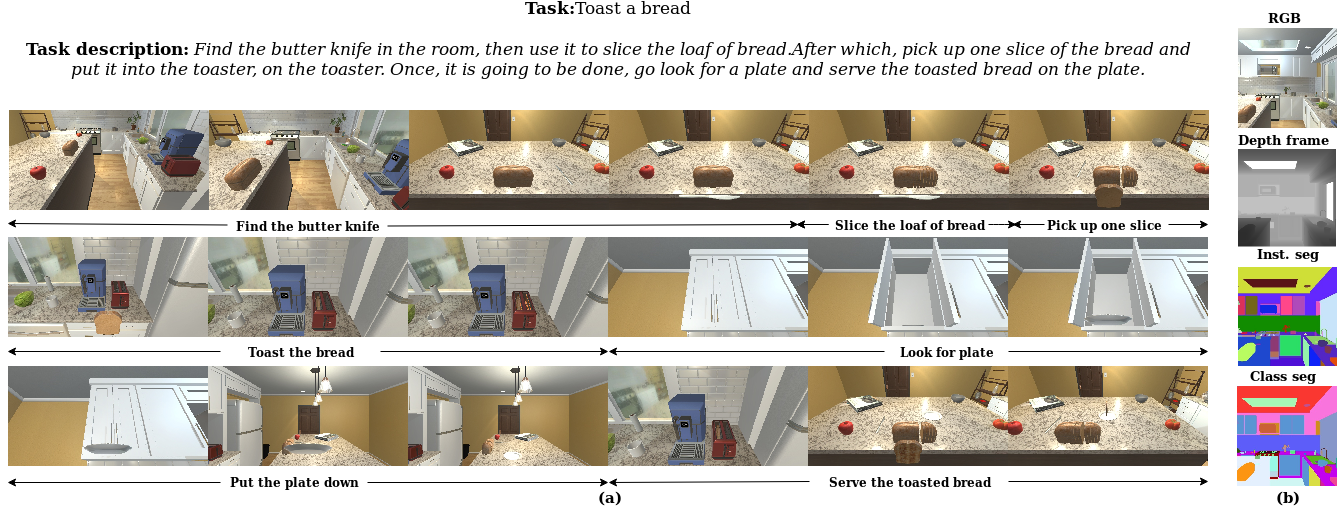}}
  \caption{Video annotated with its (a) task descriptions and atomic actions, together with (b) RGB, depth, instance segmentation and class segmentation.}
\label{fig:video}
\end{minipage}
\end{figure*}

\begin{figure}[h]
\begin{minipage}[b]{1.0\linewidth}
  \centering
  \centerline{\includegraphics[width=8.5cm]{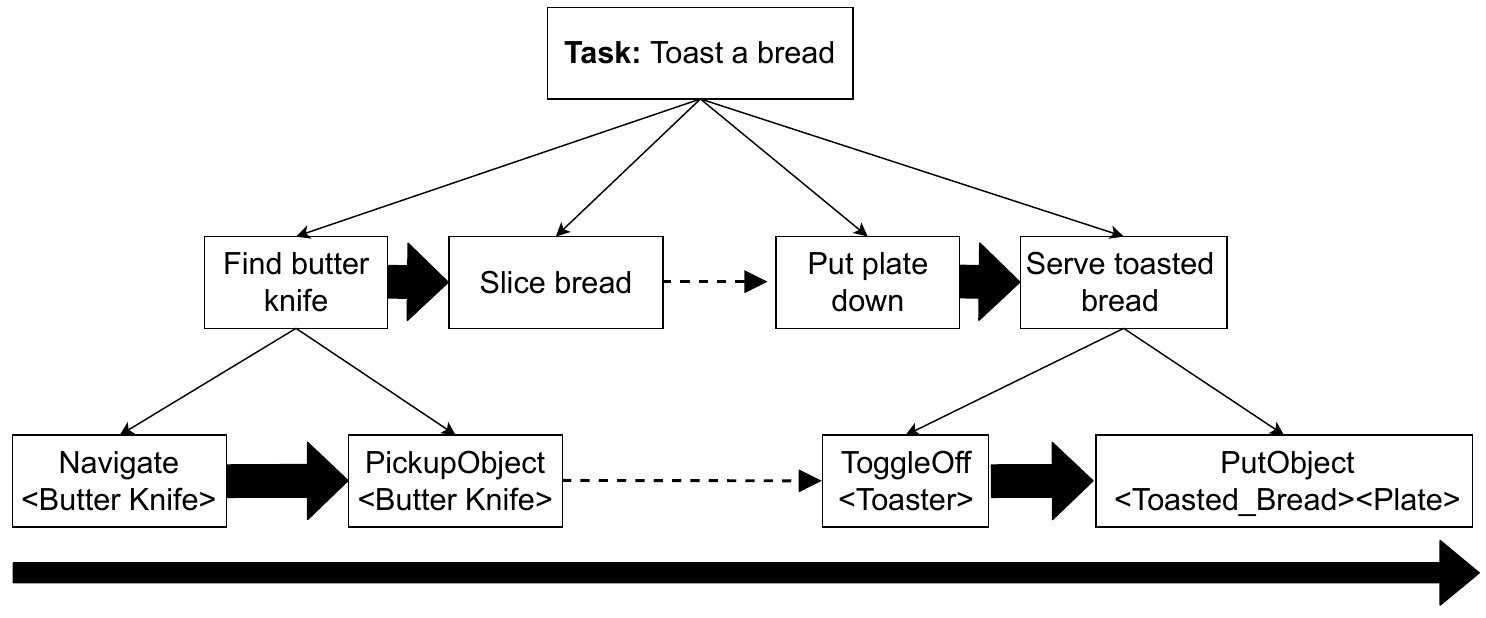}}
  \caption{Hierarchical task data structure. The first tier will be the high-level task goal as set by human, the second tier will be the task descriptions as performed by human, and the third tier will be the atomic/robotic actions.}
\label{fig:hierarchicaltaskstructure}
\end{minipage}
\end{figure}

ActioNet allows the collection and augmentation of task-based dataset in 3D environment.
For each task, annotated video and hierarchical task structure can be collected.
Referring to Fig. \ref{fig:video}, each video is annotated with its atomic action label, together with RGB, depth, instance (object) segmentation and class (object class) segmentation frames. 
Referring to Fig. \ref{fig:hierarchicaltaskstructure}, each hierarchical task structure provides task breakdown for a particular task goal.
The first tier will be the high-level task goal as set by human, the second tier will be the task descriptions as performed by human, and the third tier will be the atomic/robotic actions.
For instance, the first tier task goal of making a cup of coffee can be broken down into the second tier task descriptions of finding the mug, finding the machine, using the machine, and finally serving the coffee.
The second tier task description of serving the coffee can be further broken down into the third tier atomic/robotic actions of picking up the mug and putting down the mug. 

\begin{figure*}[thb]
\begin{minipage}[b]{1.0\linewidth}
  \centering
  \centerline{\includegraphics[width=18cm]{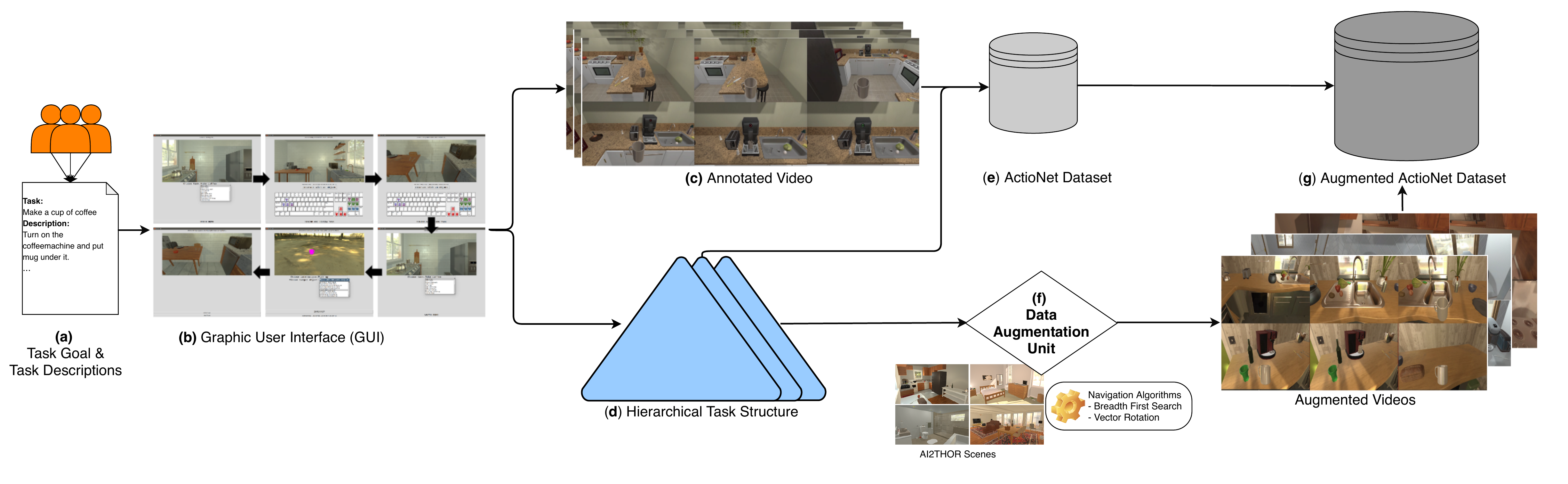}}
\caption{An overview of the ActioNet platform. 
Given an input comprising (a) task descriptions generated by annotators, using the (b) ActioNet’s GUI, the task-based dataset comprising (c) annotated videos and (d) hierarchical task structures can be collected.  
The (e) task-based dataset can be expanded using the (f) ActioNet's data augmentation unit to generate the (g) augmented dataset.}
\label{fig:platform}
\end{minipage}
\end{figure*}

An overview of ActioNet is shown in Fig. \ref{fig:platform}. 
Given the task descriptions generated by annotators, the annotated videos and hierarchical task structures can be collected using the ActioNet's GUI.
Using the hierarchical task-structures, the video dataset can be further augmented across different scenes using the data augmentation unit.
The GUI and data augmentation unit will be further elaborated in the subsequent sub-sections. 


\subsection{Graphic user interface (GUI)}
\label{subsec:gui}
Given an input task goal and corresponding task descriptions, the ActioNet's GUI can be used to collect the video and hierarchical task structure. 
The ActioNet's GUI utilizes the Tkinter library \cite{CS-R9526} which allows users to have access to the full control of all aspects of the artificial agent in the environment down to its robotic level of manipulation such as rotation and extension of its arm. 
The GUI also allows real-time object interaction capability and presents real-time update of the dynamic state changes  during users' interaction with object. 
For example, when a user drops an egg, the egg will be reflected as crack open within the GUI. 
This provides realistic environment for users to perform their given tasks, and hence capturing realistic task planning during the execution of the task in the 3D environment.



\subsection{Data augmentation unit}
\label{subsec:dataaugmentationunit}
With the hierarchical task structure collected from the ActioNet's GUI, the ActioNet's data augmentation unit can be used to augment video of a particular scene into videos of different scenes. 
This is achieved assuming that the hierarchical task structure for a certain task goal is largely consistent from scene to scene, and hence the hierarchical task structure previously collected can be used to augment the videos to different scenes.
This provides a mechanism to generate multiple videos of the same hierarchical task structure in different scenes. 

As a result of changing scene, the location of an object-of-interest might not reachable at the same location as it was in the original scene. 
Hence, the agent would need to be navigated to a position where it can successfully manipulate the object.
The shortest path between the agent and reachable location of the object is used for the navigation.
We first use the Breadth First Search algorithm \cite{Kozen1992} to map out all possible grid points that the agent can move to and then use the possible grid points to compute the shortest path between the agent and the reachable location of the object. 

For valid interactions between the agent and the object, the agent also needs to be facing the object during navigation and interaction with the object.
We use a vector rotation technique, which calculates the relationship among the agent, object, and original point of the world, to reset the agent's point of focus on the object.

\section{ActioNet dataset}
\label{sec:actionetdataset}

\subsection{Data collection}
\label{ssec:subhead}
This data collection process was done through crowd sourcing. 
First, AMT workers are provided with a task goal, the scene that they are supposed to execute the task goal and a list of items in the scene that can be utilized to achieve the task goal. 
The workers are required to provide detailed step-by-step task descriptions on how the given tasks can be achieved in the scene. 
They are required to be clear and concise in their task descriptions, as if they are generating an instructional manual. 
The task descriptions generated by the workers are based solely on their common-sense knowledge of the scene and basic human intuition in task planning.

Second, using the ActioNet's GUI, the task descriptions previously generated by the workers are used to construct the ActioNet dataset. 
Given the task goal with its corresponding task descriptions, the workers are required to use the ActioNet's GUI to control the artificial agent under different scenes in the AI2THOR environment to carry out the task as accurately as described by the task goal and task descriptions. 
Upon completion of the task, the workers will be shown video footage of the precise actions that their agent took. 
ActioNet will take the acknowledged successful entry of the task and generate the video and hierarchical task structure.

\subsection{Data analysis}
\label{ssec:subhead}
\begin{figure}[!h]
\begin{minipage}{1.0\linewidth}
  \centering
  \centerline{\includegraphics[width=8.8cm]{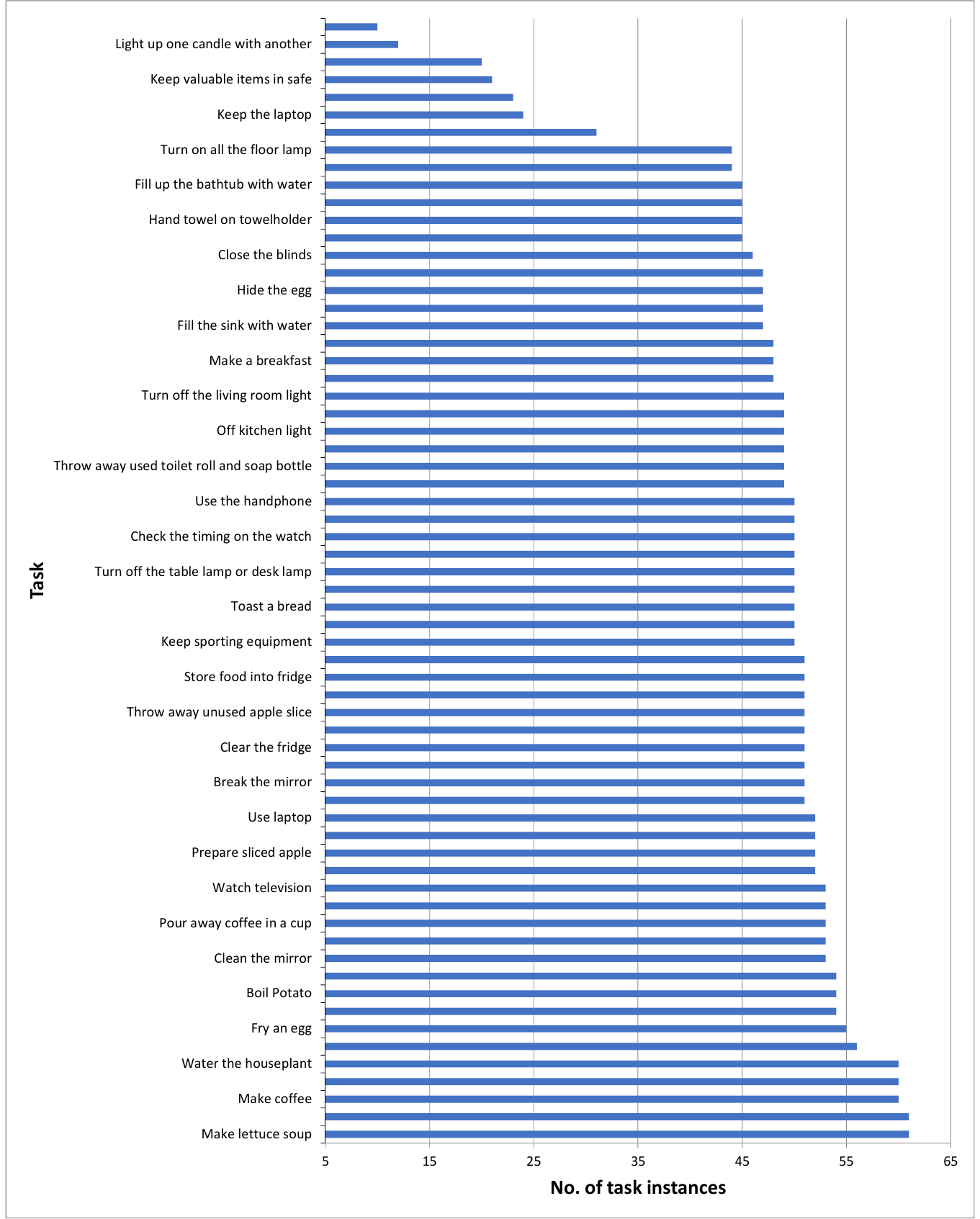}}
  \caption{Task instances per task.}
\label{fig:datasetfig}
\end{minipage}
\end{figure}

ActioNet dataset is a large-scale comprehensive task-based dataset comprising 3038 annotated videos and hierarchical task structures over 65 individual household tasks from 120 different scenes. 
Each task is annotated across three to five different scenes by 10 different annotators. 
The tasks and number of task instances for each task are illustrated in Fig. \ref{fig:datasetfig}.
The task-based dataset can be broken down into four categories, namely living room, bedroom, bathroom and kitchen.
Each category consists of 30 different room settings.
Further statistics are provided in Table \ref{tab:datanalysis}.


\begin{table}[htbp]
  \centering
  \small
  \caption{Number of tasks and task instances, as well as average number of task descriptions and atomic/robotic actions.}
\setlength{\tabcolsep}{4pt} 
    \begin{tabular}{|p{0.25\linewidth}|c|c|c|c|c|} \hline
        & Living&			& 			& 			& 	 \\ 
        & Room 	& Bedroom 	& Bathroom & Kitchen & Total \\ \hline
    No. of tasks &18  &10 &11 &26 & 65\\ \hline
    No. of task instances &794 & 404 &  526 & 1314 & 3038 \\ \hline
    Ave. no. of task descriptions &4.0 & 3.6 & 3.9 & 6.0 & 4.5 \\ \hline
    Ave. no. of atomic actions &58.7 & 60.7 & 38.3 & 93.5 & 70.5 \\ \hline
    \end{tabular}%
  \label{tab:datanalysis}%
\end{table}%

ActioNet augmented dataset is synthesized using the ActioNet data augmentation unit. 
Each data point in the 3038 task instances is augmented across 10 different scenes with five randomized object locations for each scene. 
This gives a total of over 150,000 new videos, a significant upscale of 50 times of the original video dataset.

\section{Conclusion}
\label{sec:conclusion}
In conclusion, motivated by training artificial agents to learn to generalize task planning, we have created ActioNet, the first interactive end-to-end platform for the collection and augmentation of task-based dataset in 3D environment. 
Using ActioNet, we have constructed a large-scale comprehensive task-based dataset from 65 individual household tasks over 120 rooms. 
We further up-scaled the dataset by 50 times across different scenes. 
In future, we will looking into further developing ActioNet's features and expanding on the dataset. 
The ActioNet dataset will also be used in studying various task-based learning approaches such as reinforcement learning, imitation learning, and understanding the dynamic interaction between objects in simulated environment.

\section{Acknowledgments}
This research was supported by NRF grant NRF2015-NRF-ISF001-2541 and by the Agency for Science, Technology and Research (A*STAR) under its AME Programmatic Funding Scheme (Project A18A2b0046)



\bibliographystyle{IEEEbib}
\bibliography{strings,refs}

\end{document}